\documentclass[10pt,twocolumn,letterpaper]{article}

\usepackage[pagenumbers]{cvpr}   

\usepackage{placeins}
\usepackage{algorithm}
\usepackage{algorithm}
\usepackage{algpseudocode}
\usepackage{multirow}
\usepackage{booktabs}
\usepackage{subcaption}
\usepackage{amssymb}
\usepackage{amsmath}
\usepackage{svg}
\usepackage{subcaption}

\definecolor{cvprblue}{rgb}{0.21,0.49,0.74}
\usepackage[breaklinks,colorlinks,allcolors=cvprblue]{hyperref}

\def\confName{CVPR}
\def\confYear{2027}

\title{WALDO: One-Shot Exemplar-Conditioned Object Detection in Cluttered Scenes with World-Model Features}

\author{%
\parbox{\textwidth}{\centering
Kishor Datta Gupta$^{1}$,\;
Ahmed Rafi Hasan$^{2,*}$,\;
Md. Mahfuzur Rahman$^{1,*}$,\;
Faysal Mehrab Chowdhury$^{3}$,\;
Mohd Ariful Haque$^{1}$\\[5pt]
\small
$^{1}$Dept. of Cyber Physical Systems, Clark Atlanta University, Atlanta, GA, USA\\
$^{2}$Dept. of Computer Science and Engineering, United International University, Dhaka, Bangladesh\\
$^{3}$Dept. of Computer Science and Engineering, North South University, Dhaka, Bangladesh\\[5pt]
\footnotesize\tt
kgupta@cau.edu,\;
ahasan191131@bscse.uiu.ac.bd,\;
mdmahfuzur.rahman@students.cau.edu,\\
faysal.chowdhury4@northsouth.edu,\;
mohdariful.haque@students.cau.edu\\[5pt]
\small
$^{*}$These authors contributed equally to this work.
}
}

\begin{document}
\maketitle
\begin{abstract}
Large vision-language models usually prioritize finding a particular object instance in a cluttered scene from a single reference image and indicating when it is not present. We ask whether the representations of a world model add something that the strongest self-supervised image features do not have. We introduce WALDO, a one-shot detector that finds a specific object instance in a cluttered real scene from a single reference image. In general, instance detection requires per-object training. Given one photo of an object, we want to find that exact object in a cluttered scene, or say it is not there. Open-vocabulary detectors and vision-language models find objects by category, so they cannot tell one mug from another that looks alike. We show that a small head on frozen self-supervised features can do this task. It localizes objects it has never seen, with no training per object. We then compare an image model, DINOv3, with a video world model, V-JEPA 2.1. They succeed on different queries. But V-JEPA helps only when a still image goes through its image pathway, not when the image is fed as a short video. Joining the two at the output gives a modest gain in localization. It does not hurt the model's ability to tell whether the object is there at all. Vision-language models and image-guided detectors rarely find the specific instance, and they are much slower.
\end{abstract}
\section{Introduction}
\label{sec:intro}
Consider handing a robot, a retail inventory system, or a photo search tool a
single photograph of one specific mug, not any mug of that model but
\emph{this} one, placed among near-identical items, and asking it to point
to the object or state plainly that it is absent. This task, localizing one
instance from a reference image amid clutter, underlies robotic fetch-and-place, cross-camera
re-identification and retail auditing, extending open-vocabulary retrieval
from category to identity.

Open-vocabulary detectors such as GLIP~\cite{li2022grounded},
OWL-ViT~\cite{minderer2022simple}, OWLv2~\cite{minderer2023scaling} and
Grounding DINO~\cite{liu2024grounding}, built on contrastive vision-language
models~\cite{radford2021learning,zhai2023sigmoid, jia2021scaling}, localize objects from
free-form text but answer a category-level question: a prompt returns any
matching instance, with no notion of which one was meant. Referring expression
comprehension~\cite{kazemzadeh2014referitgame,yu2018mattnet} disambiguates one
object through language alone, but every benchmark guarantees the referent is
present, so these methods are never trained to abstain. One-shot detectors
such as CoAE~\cite{hsieh2019one} and OS2D~\cite{osokin2020os2d} replace the
category label with an exemplar image and instance-level correspondence, but
drop language conditioning and treat every query as answerable.

We first tested whether this gap could be closed without training. Matching
the exemplar's pixels fails, since a separate studio photograph never
reappears pixel for pixel in the scene. Cosine similarity between frozen
backbone features often points to the right object, but how often depends
strongly on the backbone, and it gives neither a box nor a presence decision
(Supp.~D). Correspondence must be read from strong features by a trained
head, and presence must be an explicit supervised decision, not inferred post
hoc from a similarity threshold.


WALDO\footnote{Named after the search-and-find character from
\emph{Where's Waldo?}, who is likewise found by matching one distinctive
figure against a densely cluttered scene.} builds on this observation. It
trains a small detection head on frozen backbone tokens and correlates every
scene patch with every object token of the exemplar, in the backbone's own
feature space. FiLM conditioning~\cite{perez2018film} from the exemplar
follows, and a dedicated presence head makes absence a
trained output rather than a byproduct of low confidence. One head of about
3.3M parameters is trained per backbone, and no backbone is fine-tuned.
Heads on DINOv3~\cite{simeoni2025dinov3} and V-JEPA~2.1~\cite{mur2026v}
localize different queries, so WALDO combines them by late fusion: presence
comes almost entirely from DINOv3, with the weight chosen on validation
objects, and boxes come from both heads by weighted box fusion with fixed,
equal weights. The fusion improves top-1 localization only when V-JEPA~2.1
is read through its image pathway instead of as a two-frame clip.

\textbf{Contributions:}
\begin{enumerate}
    \item[(i)] We formulate exemplar-conditioned instance detection with explicit
    presence prediction, and evaluate it on novel objects with cross-view
    references.
    \item[(ii)] We propose WALDO, a lightweight head on frozen backbones with dense
    exemplar correlation and a trained presence output, and show that late fusion
    of its DINOv3 and V-JEPA~2.1 heads localizes unseen instances slightly
    better than DINOv3 alone, by a margin close to run-to-run variation, without
    lowering presence AUROC, while concat and gated fusion
    learned inside one head do not.
    \item[(iii)] We build training episodes for novel
    instances from studio captures alone, drawing the present/absent label after
    the query region so that region geometry carries no label information.
    \item[(iv)] We show by ablation that the head uses the exemplar's identity, and that
    an image-guided open-vocabulary detector, a vision-language model and template
    matching almost never localize the specific instance.
\end{enumerate}

\section{Related Work}
\label{sec:related_work}
Few-shot object detection supplies several support images per class and asks
whether a query region belongs to that class~\cite{kang2019few, wang2020frustratingly}; despite
needing less supervision per class this remains a category-level question, not
the single-instance question CoAE~\cite{hsieh2019one} and
OS2D~\cite{osokin2020os2d} pose. More recent methods build single-instance
correspondence on frozen self-supervised backbones rather than learning it from
scratch: DE-ViT projects DINOv2 features into a prototype
subspace~\cite{zhang2023detect} and PerSAM derives an attention prior from
SAM's image encoder~\cite{zhang2024personalize}, both computed once from a
single exemplar. A single prototype or prior averages views of the object that
need not match the scene, whereas dense query--support correlation, as in
OS2D, hypercorrelation for few-shot segmentation~\cite{min2021hypercorrelation}
and Matcher~\cite{liu2024matcher}, compares every scene location with every
part of the exemplar. Instance detection benchmarks such as
InsDet~\cite{shen2023high} provide multi-view captures of each object, and
NIDS-Net~\cite{lu2025adapting} and IDOW~\cite{shen2025solving} match external
proposals against embeddings of these views, so both need several reference
views and a separate proposal stage. Visual-prompt detectors such as
T-Rex2~\cite{jiang2024trex2} and image-guided OWLv2~\cite{minderer2023scaling}
accept an exemplar in place of text, and vision-language models such as
Qwen2.5-VL~\cite{bai2025qwen25vltechnicalreport} can be shown a reference image, but
\cref{sec:results} shows that they rarely localize the specific instance.

Open-set recognition originates in classification, where Scheirer et al.\
formalized rejecting classes never seen at training
time~\cite{scheirer2012toward}. Open-set detection brings this to localization:
Dhamija et al.\ formalized the first protocol~\cite{dhamija2020overlooked}, ORE
incrementally learns categories flagged as unknown~\cite{joseph2021towards},
and Han et al.\ push unknown instances into low-density regions of the feature
space~\cite{han2022expanding}. All treat absence as \emph{category novelty},
with no metric for instance-level absence: whether one specific, previously
seen object rather than a same-category look-alike is present in a new image.
Generalized referring expression benchmarks add expressions with no
target~\cite{liu2023gres, he2023grec}, but the referent is described in text,
not shown by an exemplar of one instance. The distinction between instance and
category is easy to lose, since shortcut learning lets cues such as region
size predict a label without the exemplar ever being
consulted~\cite{geirhos2020shortcut}.

Video world models learn representations by predicting the features of masked
or future content rather than its pixels. V-JEPA~\cite{bardes2024revisiting}
and V-JEPA~2~\cite{assran2025vjepa2} predict masked spatio-temporal regions in
latent space, and V-JEPA~2.1 adds dense features and a separate image
tokenizer for still images~\cite{mur2026v}. These models are evaluated
on video understanding, anticipation, planning and dense prediction. Whether
their features carry instance identity that image self-supervision misses, and
whether a still image should be read as a repeated clip or through the image
pathway, has not been tested.

Frozen vision foundation models differ in what their features capture, which
raises the question of how to combine them. AM-RADIO distills several
teachers (CLIP, DINOv2 and SAM) into a single student
encoder~\cite{ranzinger2024amradio}, and its latest release,
C-RADIOv4~\cite{ranzinger2026c}, distills SigLIP2, DINOv3 and
SAM3~\cite{carion2026sam}. The Perception
Encoder shows that the most general features of a contrastive
vision-language model lie in its intermediate layers rather than at its
output, and aligns them for dense prediction in
PE-Spatial~\cite{bolya2026perception}. Both are evaluated on classification,
segmentation, depth and multimodal benchmarks, not on localizing one specific
instance and deciding whether it is present. The alternative to distillation
keeps the models separate and combines their predictions, as weighted box
fusion does for ensembles of detectors~\cite{solovyev2021weighted}.

Like one-shot detectors, WALDO resolves instance rather than category identity
from a single exemplar view, and unlike instance detectors it needs neither
multiple reference views nor external proposals. Unlike referring expression
methods, it does not assume the referent is present: presence is a separately
supervised output. Unlike agglomerative distillation, it keeps the frozen
backbones and their heads separate and combines them only at the output, which
on InsDet outperforms concat and gated fusion learned inside one head
(\cref{sec:res-fusion}).
\section{Method}
\label{sec:method}

\begin{figure*}[t]
\centering
\includegraphics[
    width=1\textwidth,
    height=1.2\textheight,
    keepaspectratio
]{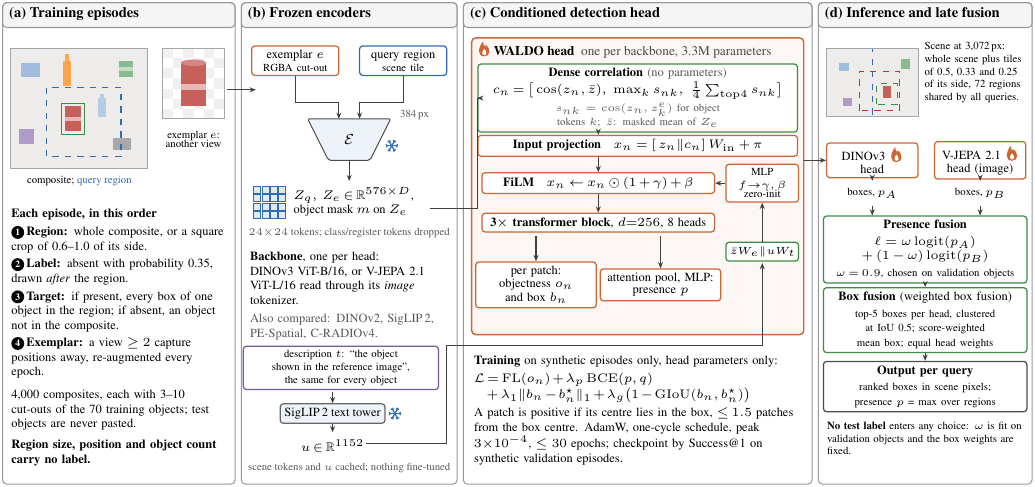}
\caption{\textbf{WALDO.} (1)~Training episodes on synthetic composites. The
exemplar is a different view of the queried object, and the present/absent
label is drawn after the query region. (2)~A frozen backbone encodes the
exemplar and the query region on a $24\times24$ grid; a frozen SigLIP~2 text
tower encodes the description. (3)~The trainable head correlates every query
patch with the exemplar's object tokens, applies FiLM and three transformer
blocks, and predicts per-patch boxes and objectness and a presence probability.
(4)~At test time each head runs on multi-scale tiles. The DINOv3 and
V-JEPA~2.1 heads are fused at the output: presence comes almost entirely from
DINOv3, boxes from both heads.}
\label{fig:architecture}
\end{figure*}

WALDO (\cref{fig:architecture}) trains a small conditioned head on a frozen
image backbone; no encoder
is fine-tuned and no test object is seen in training. Training scenes are
synthesized from studio captures (\cref{sec:episodes}). The exemplar and the
scene show different viewpoints, so the head compares every scene patch with
every part of the exemplar (\cref{sec:head,sec:training}). Objects are small,
so inference runs on multi-scale tiles (\cref{sec:inference}). Backbones
localize different queries, so two are fused at the output (\cref{sec:fusion}).

\subsection{Problem setting}
\label{sec:problem}

A query is a triple $(I, e, t)$: a scene image $I$, an exemplar $e$ and a
description $t$. The exemplar is one photograph of a specific object instance
with its foreground mask, taken from its own viewpoint against its own
background, so the object cannot be found by matching its exact pixels. The
model returns ranked boxes $\{(b_k, s_k)\}_{k=1}^{K}$ and a presence
probability $p \in [0,1]$. Every scene is queried with every object of the
evaluation set; a query is \emph{present} if the object is annotated in the
scene and \emph{absent} otherwise, so absence is a property of the exemplar,
not of the scene. Objects are split into disjoint training, validation and test
sets, so a test query asks about an unseen instance, from one view, without
per-object fine-tuning.

\subsection{Training episodes from object captures}
\label{sec:episodes}

InsDet~\cite{shen2023high} provides $24$ masked studio views per object, real
cluttered test scenes and object-free background photographs, but no training
scenes. We synthesize them by cut-and-paste~\cite{dwibedi2017cut}.

\paragraph{Composites.}
Each composite pastes $3$ to $10$ masked cut-outs onto a random square
background crop resized to $768\times768$. An object's long side is drawn
log-uniformly between $0.044$ and $0.479$ of the canvas side, the 5th to 95th
percentile of real object sizes relative to the test-time tiles
(\cref{sec:inference}). Pastes that would cover more than $30\%$ of an earlier
object are rejected. Boxes cover only \emph{visible} pixels, and an object
left less than $60\%$ visible is painted over with background, so no visible
object is labelled absent. We generate $4{,}000$ training composites from
training objects and $400$ validation composites queried only about
validation objects.

\paragraph{Episodes.}
An episode is built in a fixed order. (1)~\emph{Region}: the whole composite
(probability $0.5$) or a square crop of $0.6$ to $1.0$ of its side, redrawn
until at least one query object lies $90\%$ inside it. (2)~\emph{Label}:
absent with probability $0.35$. (3)~\emph{Target}: a present episode targets
every box of one object inside the region; an absent episode queries an object
that appears nowhere in the composite. (4)~\emph{Exemplar}: a cut-out of the
queried object from a view at least two capture positions away from the pasted
one. Because the region is drawn before the label and always contains a
queryable object, its size, position and object count are independent of the
label. With a first sampler that made every object-free region absent, object
count alone predicted the label with AUROC $0.674$; a unit test now requires
each of these quantities to stay within $0.12$ of chance.

\paragraph{Exemplars and text.}
Exemplars are re-augmented every epoch with rotation, colour jitter and blur
(Supp.~G) and padded to a square with transparent pixels. The description is
the same sentence for every object, ``the object shown in the reference
image'', so localization must come from the exemplar.

\subsection{Conditioned detection head}
\label{sec:head}

\paragraph{Frozen features.}
The query region is resized to $384\times384$ and encoded by a frozen ViT into
$N = 24\times24$ patch tokens $z_n \in \mathbb{R}^{D}$. The exemplar is encoded the same way into tokens
$z^{e}_k$, and $m_k = 1$ marks token $k$ as object when at least half of its
patch is opaque. The pooled exemplar is
$\bar{z} = \sum_k m_k z^{e}_k / \sum_k m_k$, and the frozen SigLIP~2 text
tower~\cite{tschannen2025siglip} encodes the description into
$u \in \mathbb{R}^{1152}$. The head reads a backbone only through its token
grid and width $D$.

\paragraph{Dense correlation.}
Across viewpoints, different parts of the object are visible in the exemplar
and in the scene, and a pooled vector averages them together. The head
therefore compares each scene patch with every object token, in the backbone's
own feature space. With $s_{nk} = \cos(z_n, z^{e}_k)$ over object tokens $k$,
patch $n$ receives
\begin{equation}
c_n = \Big[\cos(z_n, \bar{z}),\;\; \max_{k} s_{nk},\;\;
      \tfrac{1}{4}\textstyle\sum_{k \in \mathrm{top}_4(n)} s_{nk}\Big],
\label{eq:corr}
\end{equation}
where $\mathrm{top}_4(n)$ are its four most similar object tokens: does the
patch resemble the whole object, some part of it, or several parts? The
features are appended to the tokens before the input projection,
\begin{equation}
x_n = \big[\,z_n \,\Vert\, c_n\,\big]\, W_{\text{in}} + \pi,
\qquad W_{\text{in}} \in \mathbb{R}^{(D+3)\times d},
\label{eq:proj}
\end{equation}
with $d = 256$ and a learned bias $\pi$. The head has no positional
embedding; position enters only through box decoding (\cref{eq:box}).

\paragraph{Conditioning.}
The pooled exemplar and the description are projected to $d$ dimensions,
concatenated, and mapped by a two-layer MLP $f$ to a FiLM~\cite{perez2018film}
modulation of every patch,
\begin{equation}
[\gamma; \beta] = f\big([\,\bar{z} W_e \,\Vert\, u W_t\,]\big),
\qquad
x_n \leftarrow x_n \odot (1 + \gamma) + \beta .
\label{eq:film}
\end{equation}
The last layer of $f$ is zero-initialized, so FiLM starts as the identity.
FiLM is the only path from the description to the prediction.

\paragraph{Prediction.}
Three pre-norm transformer blocks ($8$ heads) and a layer norm follow. Linear
layers read out, per patch, an objectness logit $o_n$ and box parameters
$r_n \in \mathbb{R}^4$, decoded relative to the patch centre $\mu_n$ in
region-normalized coordinates:
\begin{equation}
(b^{x}_n, b^{y}_n) = \mu_n + \tfrac{2}{G}\big(\sigma(r_n^{1:2}) - \tfrac{1}{2}\big),
\qquad
(b^{w}_n, b^{h}_n) = \sigma(r_n^{3:4}),
\label{eq:box}
\end{equation}
with grid side $G = 24$ and logistic $\sigma$. The centre moves at most one
patch, so objectness and box refer to the same location. Presence is predicted
by an MLP from an attention-pooled summary
$\bar{x} = \sum_n \mathrm{softmax}_n(a^{\top} x_n)\, x_n$: absence means that
\emph{no} patch matches, and a mean over $576$ patches would let the background
outweigh the one patch that found the object. The head has $3.26$M to $3.46$M
parameters; only the input projections depend on $D$.

\subsection{Training}
\label{sec:training}

Following the centre sampling of FCOS~\cite{tian2019fcos}, a patch is positive
for a target box if its centre lies inside the box and within $1.5$ patch
widths of the box centre; every box keeps at least its nearest patch, and the
smaller box wins a shared patch. With positive patches $\mathcal{P}$,
$n_{\text{pos}} = \max(1,|\mathcal{P}|)$, objectness targets $y_n$, assigned
boxes $b^{\star}_n$ and presence label $q$, the loss is
\begin{equation}
\begin{split}
\mathcal{L} = {}& \frac{1}{n_{\text{pos}}}\sum_{n} \mathrm{FL}(o_n, y_n)
  + \lambda_p\, \mathrm{BCE}(p, q) \\
  & + \frac{1}{n_{\text{pos}}}\sum_{n \in \mathcal{P}}
  \Big[ \tfrac{\lambda_1}{4} \lVert b_n - b^{\star}_n \rVert_1
  + \lambda_g \big(1 - \mathrm{GIoU}(b_n, b^{\star}_n)\big) \Big],
\end{split}
\label{eq:loss}
\end{equation}
with the focal loss~\cite{lin2017focal} FL, the generalized
IoU~\cite{rezatofighi2019generalized}, $\lambda_1 = 5$, $\lambda_g = 2$ and
$\lambda_p = 1$. Absent episodes contribute only the objectness and presence
terms. We train with AdamW~\cite{loshchilov2017decoupled} under a one-cycle
schedule (peak learning rate $3\times10^{-4}$, batch size $8$) for at most
$30$ epochs (Supp.~G). The checkpoint is selected by Success@1
on the synthetic validation episodes, with presence AUROC as tie-break; real
scenes are never used for selection.

\subsection{Inference on high-resolution scenes}
\label{sec:inference}

InsDet objects are small: the median object's long side is $5.6\%$ of the
scene's short side, about $1.4$ patches if the scene is encoded once. Each
scene is resized to a long side of $3{,}072$\,px and covered at four scales,
the whole scene and square tiles of $0.5$, $0.33$ and $0.25$ of the short side
with at least $25\%$ overlap. Each of the resulting $72$ regions is encoded
once at $384\times384$, and its tokens are shared by every query on the scene.
The exemplar is one unaugmented view (view $0$). Boxes from all regions are
mapped to scene coordinates; the $300$ highest by objectness are kept, and
non-maximum suppression at IoU $0.5$ returns at most $100$. Scene presence is
the maximum over regions. With several exemplar views, boxes are pooled and
presence is the maximum over views and regions.

\subsection{Fusing two frozen backbones}
\label{sec:fusion}

Each backbone gets its own head, trained independently with the recipe above,
and the heads are combined only at their outputs.

\paragraph{Reading V-JEPA~2.1 as an image model.}
V-JEPA~2.1~\cite{mur2026v} is a video model whose video tokenizer groups frames
into two-frame tubelets, so a still image is usually fed as a repeated
two-frame clip. The model also has a separate image tokenizer, used in
pretraining for single images. We feed the scene and the exemplar through this
image pathway and read the last block's tokens.

\paragraph{Late fusion.}
Presence and boxes are fused separately, because a head can be reliable at one
and not the other. For heads $A$ and $B$, fused presence is a weighted average
of logits,
\begin{equation}
\ell = \omega\, \mathrm{logit}(p_A) + (1 - \omega)\, \mathrm{logit}(p_B),
\label{eq:fuse}
\end{equation}
where $\omega \in \{0, 0.05, \dots, 1\}$ maximizes presence AUROC on the
\emph{validation} objects queried on the real scenes (ties broken towards
$0.5$), so no test label enters the choice. Boxes are fused by weighted box
fusion~\cite{solovyev2021weighted} over each head's five highest-scoring boxes,
with equal head weights and no tuned parameter: boxes are clustered greedily at
IoU $0.5$, a cluster's box is the score-weighted mean of its members, and its
score sums the best score each head contributed. Agreement between the heads
therefore raises a box, and one head cannot raise it with duplicates. The main
model fuses DINOv3~\cite{simeoni2025dinov3} ($A$) with V-JEPA~2.1 read through
its image pathway ($B$). Validation selects $\omega = 0.9$, so presence is
almost entirely DINOv3's and V-JEPA~2.1 contributes through the boxes.

\paragraph{Learned fusion.}
As a control, two heads learn the combination from both backbones' tokens on
the same grid, with \cref{eq:corr} computed per backbone. \emph{Concat fusion}
maps both token sets and correlation features through one input projection.
\emph{Gated fusion} projects each backbone separately and mixes them per patch
with softmax weights predicted from the tokens and the conditioning vector;
the gate starts uniform, and one backbone is masked in $15\%$ of training
episodes so that the gate does not collapse onto the stronger one.

\section{Experimental Setup}
\label{sec:experiments}

\subsection{Benchmark and protocol}
\label{sec:exp-data}

InsDet~\cite{shen2023high} contains $100$ objects, each with $24$ masked
studio views, and $160$ real cluttered scenes ($120$ easy, $40$ hard;
$8{,}192\times6{,}144$\,px) with $3{,}081$ annotated boxes. We split the
objects, not the scenes, at seed~$0$ into $70$ training, $10$ validation and
$20$ test objects (Supp.~G). No real scene is seen in training. Every test object is queried in every scene
with view~$0$ as the exemplar: $3{,}200$ queries, $704$ present and $2{,}496$
($78\%$) absent. The $10$ validation objects are queried the same way
($1{,}600$ queries) and only choose the fusion presence weight, never a
checkpoint.

\paragraph{Closed-set reference.}
To measure the cost of novelty, we also train heads on composites of all $100$
objects and score them on the same $20$ test objects, now seen in training.
Published InsDet detectors~\cite{shen2023high,lu2025adapting} use multiple
reference views and external proposals in this closed setting, so we do not
compare our one-view, novel-object AP with theirs. IDOW~\cite{shen2025solving}
also reports novel instances, but with multiple views, external proposals and
its own split, so its numbers are not comparable either.

\begin{table}[t]
\centering\footnotesize
\caption{\textbf{Frozen backbones.} Every encoder emits a $24\times24$ token
grid, so the input side is $24\times$ the patch size; only the head's input
projections depend on the token width $D$.}
\label{tab:setup}
\setlength{\tabcolsep}{2.5pt}
\begin{tabular}{l l cccc}
\toprule
Backbone & Pretraining & Patch & Input & $D$ & Head \\
\midrule
DINOv3 ViT-B/16          & self-sup.  & 16 & 384 & 768  & 3.26M \\
DINOv2 ViT-B/14          & self-sup.  & 14 & 336 & 768  & 3.26M \\
PE-Spatial B/16          & image-text & 16 & 384 & 768  & 3.26M \\
SigLIP~2 SO400M/16       & image-text & 16 & 384 & 1152 & 3.46M \\
C-RADIOv4 SO400M         & distilled  & 16 & 384 & 1152 & 3.46M \\
V-JEPA~2.1 ViT-L/16      & video      & 16 & 384 & 1024 & 3.39M \\
\bottomrule
\end{tabular}
\end{table}

\subsection{Backbones, fusion, ablations and baselines}
\label{sec:exp-arms}

Six frozen backbones (\cref{tab:setup}) share the head, episodes, loss,
optimizer, split and seed, so the representation is the only free variable:
the self-supervised DINOv3~\cite{simeoni2025dinov3} and
DINOv2~\cite{oquab2023dinov2}; the image-text SigLIP~2~\cite{tschannen2025siglip}
and PE-Spatial~\cite{bolya2026perception}, tuned for dense features;
C-RADIOv4~\cite{ranzinger2026c}, distilled from several teachers; and
V-JEPA~2.1~\cite{mur2026v}, a video world model trained by feature prediction,
without its predictor. The last block's patch tokens are read throughout. The
token grid, not the pixel count, is matched. Encoder size is not (ViT-B to
ViT-L), so the comparison ranks representations as released, not objectives
at equal capacity.

Three controls test what the fusion gain depends on: V-JEPA~2.1 read as a
two-frame clip, other partners (PE-Spatial, C-RADIOv4, DINOv2), and concat and
gated fusion of the same pairs. The DINOv2 head was not run on the validation
objects, so its fusion uses $\omega = 0.5$. For the main pair, each half of the rule is
also replaced: non-maximum suppression over the pooled boxes, with raw or
z-scored scores, for weighted box fusion; and the mean probability, mean rank,
maximum or equal-weight logit for the validation-weighted logit.

Inference-time ablations reuse a checkpoint and change one input:
\emph{no correlation} zeroes the features of \cref{eq:corr}, \emph{zero text}
zeroes $u$, \emph{swapped exemplar} conditions on another test object's
exemplar while scoring against the original object, and \emph{4-view exemplar}
pools views $0$, $6$, $12$ and $18$. Retrained variants pool the exemplar
instead of correlating densely, add a cross-attention layer from the patches
to the exemplar tokens after FiLM, or pass the object's name as the
description. PE-Spatial and C-RADIOv4 are pretrained at $512$\,px, so two more
variants encode them at $512$\,px with the grid pooled to $24\times24$, or read
four blocks instead of the last.

\paragraph{Baselines.}
All baselines answer the same queries with the same exemplar, without
fine-tuning, and are scored by the same code. \textbf{Template matching} slides
the grayscale exemplar over a six-level scale pyramid with normalized
cross-correlation; presence is the best correlation.
\textbf{OWLv2}~\cite{minderer2023scaling}
(\texttt{owlv2-base-patch16-ensemble}) runs image-guided detection on the same
$72$ regions, with raw similarities so that scores are comparable across
scenes; presence is the maximum similarity.
\textbf{Qwen2.5-VL-7B}~\cite{bai2025qwen25vltechnicalreport} is shown the
exemplar and the scene and asked whether the exact object appears, scored as
$P(\text{yes})$, and for its box as JSON.

\subsection{Metrics and statistics}
\label{sec:exp-metrics}

\emph{Success@$K$} (S@$K$) is the fraction of present queries with a box of
IoU $\geq 0.5$ in the top $K$; S@1 is the primary metric, since a deployed
system acts on its first answer. \emph{Presence AUROC} ranks all $3{,}200$
queries by $p$. mAP, AP$_{50}$ and AP$_{75}$ are COCO-style AP per object,
averaged over the $20$ test objects; we omit AP$_S$, since only $7$ of the
$705$ test instances are small (one present query has two instances). The fused model keeps five boxes per head, so its AP is a lower
bound. Intervals are 95\% percentile intervals over $1{,}000$ bootstrap
resamples of scenes. Differences use a paired bootstrap on the same resamples
and are significant only when the interval excludes zero. Each arm is trained
once, at seed~$0$. The bootstrap captures scene sampling, not training
noise: retraining the closed-set DINOv3 head with the same configuration moved
S@1 over all $100$ objects by $0.029$ (paired $[-0.042, -0.015]$; $0.024$ on
the $20$ test objects, Supp.~A), so we do not attribute smaller S@1
differences to a model change. No choice uses a test
label. Everything runs in PyTorch on one NVIDIA RTX 6000 Ada GPU (48\,GB);
training a head takes $68$ to $247$ minutes.

\section{Results}
\label{sec:results}

\begin{table*}[t]
\centering\small
\caption{\textbf{Instance localization and presence on InsDet novel objects}
($20$ unseen objects, $160$ real scenes, $3{,}200$ queries, one exemplar view).
Every WALDO row uses the same head. Of the late-fusion partners, only
V-JEPA~2.1 read through its image pathway raises DINOv3's S@1 reliably, and it
leaves AUROC unchanged. Presence weights on DINOv3, chosen on validation
objects: $0.90$, $0.75$, $0.75$, $0.80$. $^*$Paired 95\% interval of the S@1
difference against DINOv3 excludes zero. $^\dagger$From each head's top five
boxes, a lower bound (DINOv3: $0.128$ mAP). $^\ddagger$No validation run; equal
presence weight, so its AUROC is not comparable. \textbf{Bold}: best;
\underline{underlined}: second best.}
\label{tab:main}
\setlength{\tabcolsep}{5pt}
\begin{tabular}{l cc cc ccc}
\toprule
Method & S@1$\uparrow$ & 95\% CI & S@5$\uparrow$ & AUROC$\uparrow$ & mAP$\uparrow$ & AP$_{50}\uparrow$ & AP$_{75}\uparrow$ \\
\midrule
\multicolumn{8}{l}{\emph{Baselines, not trained on InsDet}} \\
Template matching (NCC) & .006 & [.001, .012] & .028 & .525 & .000 & .000 & .000 \\
OWLv2, image-guided     & .003 & [.000, .007] & .003 & .563 & .000 & .002 & .000 \\
Qwen2.5-VL-7B           & .006 & [.001, .012] & .006 & .804 & .000 & .001 & .000 \\
\midrule
\multicolumn{8}{l}{\emph{WALDO head on one frozen backbone}} \\
DINOv3 ViT-B/16             & .446 & [.411, .483] & .741 & \underline{.872} & .129 & .299 & .088 \\
V-JEPA~2.1, image           & .332 & [.296, .368] & .675 & .714 & .065 & .143 & .045 \\
V-JEPA~2.1, two-frame video & .290 & [.255, .327] & .634 & .742 & .051 & .120 & .035 \\
C-RADIOv4 SO400M            & .254 & [.221, .289] & .572 & .681 & .059 & .139 & .044 \\
DINOv2 ViT-B/14             & .237 & [.206, .271] & .500 & .707 & .029 & .090 & .009 \\
PE-Spatial B/16             & .210 & [.179, .238] & .496 & .639 & .021 & .071 & .005 \\
SigLIP~2 SO400M/16          & .196 & [.165, .232] & .491 & .668 & .036 & .106 & .011 \\
\midrule
\multicolumn{8}{l}{\emph{Late fusion with DINOv3, training-free (\cref{sec:fusion})}} \\
+ V-JEPA~2.1, image (WALDO)    & \textbf{.479}$^*$ & [.443, .517] & \textbf{.771} & \underline{.872} & \textbf{.154}$^\dagger$ & \textbf{.322}$^\dagger$ & \textbf{.125}$^\dagger$ \\
+ V-JEPA~2.1, two-frame video  & .443 & [.408, .481] & .754 & \textbf{.876} & .134$^\dagger$ & .298$^\dagger$ & .097$^\dagger$ \\
+ PE-Spatial                   & \underline{.457} & [.424, .495] & \underline{.759} & .871 & .134$^\dagger$ & .307$^\dagger$ & .087$^\dagger$ \\
+ C-RADIOv4                    & .442 & [.404, .482] & .739 & .869 & \underline{.148}$^\dagger$ & \underline{.315}$^\dagger$ & \underline{.106}$^\dagger$ \\
+ DINOv2$^\ddagger$            & .412$^*$ & [.376, .449] & .713 & .831 & .109$^\dagger$ & .260$^\dagger$ & .072$^\dagger$ \\
\bottomrule
\end{tabular}
\end{table*}

\subsection{Backbones and baselines}
\label{sec:res-backbones}

Under one head, the backbone moves S@1 from $0.196$ (SigLIP~2) to $0.446$
(DINOv3). DINOv3 leads every single-backbone metric in \cref{tab:main} and both
scene difficulties in \cref{tab:errors}, and its S@1 interval does not overlap
that of the next backbone, V-JEPA~2.1 read through its image pathway
($0.332$). Reading V-JEPA~2.1 through this pathway instead of as a clip raises
its S@1 by $0.043$ ($[+0.015, +0.071]$) but lowers its AUROC by $0.028$. The
objective alone does not set the order: the self-supervised DINOv2 reaches
$0.237$.

The gap is specific to novel objects. Trained on all $100$ objects and scored
on the same $20$, PE-Spatial reaches $0.501$ and C-RADIOv4 $0.528$, against
$0.534$ and $0.510$ for two DINOv3 runs (Supp.~A): novelty costs DINOv3 $0.06$
to $0.09$ S@1 and the other two about $0.28$. Encoding PE-Spatial and
C-RADIOv4 at $512$\,px or reading four blocks does not close the gap
(\cref{tab:ablation}a), although $512$\,px raises their AUROC from $0.639$
to $0.721$ and from $0.681$ to $0.753$.

No baseline localizes the instance (S@1 $\leq 0.006$). Qwen2.5-VL-7B ranks
presence with AUROC $0.804$,
but its boxes almost never reach IoU $0.5$.

\begin{figure}[t]
\centering
\includegraphics[width=\columnwidth]{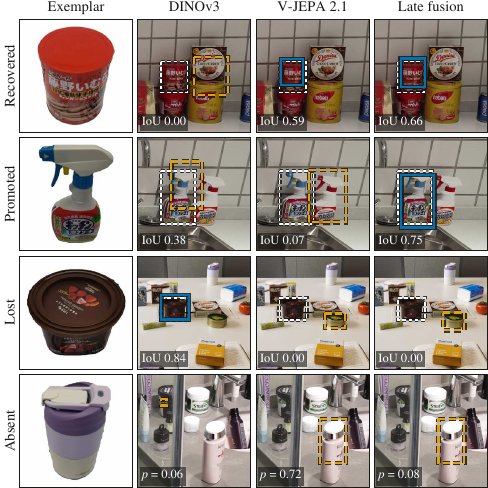}
\caption{\textbf{Late fusion changes answers in both directions.} One test
query per row; exemplar: view~$0$ cut-out. White dashed: ground truth; blue:
correct top-1 box (IoU $\geq 0.5$); orange dashed: wrong one.
\emph{Recovered} ($42$ queries): DINOv3 picks a neighbouring product;
V-JEPA~2.1 and the fused box pick the right one. \emph{Promoted} ($29$):
neither head's top-1 box is correct, but their fused box is. \emph{Lost}
($46$ of the $48$ lost hits; in the other two both heads are right): DINOv3 is
right, V-JEPA~2.1 is wrong, and so is the fused box, here V-JEPA~2.1's
similar-looking container. \emph{Absent}: V-JEPA~2.1 reports a
similar bottle as present ($p$: presence probability); fused presence follows
DINOv3 and stays low.}
\label{fig:fusion}
\end{figure}

\begin{table*}[t]
\centering\footnotesize
\newcommand{\gry}[1]{\textcolor{gray}{#1}}
\caption{\textbf{Ablations} on InsDet novel objects. $^*$Paired 95\% interval
excludes zero; --: not run. (a)~Swapping the exemplar collapses every
backbone, and removing correlation collapses DINOv3; no variant brings
PE-Spatial (PE) or C-RADIOv4 (RADIO) close to DINOv3. $\Delta$ and $^*$ are
against each backbone's default head; the last two variants apply only to PE
and RADIO. (b)~Each half of the fusion rule changed alone ($\omega$: presence
weight on DINOv3), $\Delta$ against DINOv3 alone. S@1 depends only on the boxes
and AUROC only on presence, so grey cells repeat the first row. (c)~Heads
trained on both backbones, each expert of the gated head alone, and late
fusion.}
\label{tab:ablation}
\setlength{\tabcolsep}{3pt}
\begin{minipage}[t]{0.52\textwidth}\centering
\textbf{(a)} What the head uses\\[2pt]
\begin{tabular}{l ccc cc}
\toprule
 & \multicolumn{3}{c}{DINOv3} & \multicolumn{2}{c}{S@1} \\
\cmidrule(lr){2-4}\cmidrule(lr){5-6}
Variant & S@1 & $\Delta$S@1 & AUROC & PE & RADIO \\
\midrule
Default (dense correlation) & .446 & ref. & .872 & .210 & .254 \\
\midrule
\multicolumn{6}{l}{\emph{Same checkpoint, one input changed}} \\
No correlation      & .024 & $-.422^*$ & .601 & .206 & .241 \\
Swapped exemplar    & .054 & $-.392^*$ & .554 & .020$^*$ & .021$^*$ \\
Zero text           & .446 & $\phantom{-}.000$ & .872 & .210 & .256 \\
4-view exemplar     & .440 & $-.006$   & .875 & .205 & .251 \\
\midrule
\multicolumn{6}{l}{\emph{Head retrained}} \\
Pooled exemplar     & .244 & $-.202^*$ & .756 & .185 & .209$^*$ \\
Dense + cross-attention & .253 & $-.193^*$ & .716 & .197 & .244 \\
Object name as text & --   & --        & --   & .250$^*$ & .266 \\
$512$\,px, pooled   & --   & --        & --   & .180 & .266 \\
Four blocks         & --   & --        & --   & .141$^*$ & .189$^*$ \\
\bottomrule
\end{tabular}
\end{minipage}\hfill
\begin{minipage}[t]{0.46\textwidth}\centering
\textbf{(b)} Fusion rule, DINOv3 + V-JEPA~2.1 (image)\\[2pt]
\begin{tabular}{l cc cc}
\toprule
Rule & S@1 & $\Delta$S@1 & AUROC & $\Delta$AUROC \\
\midrule
Ours: WBF; logit, $\omega=0.9$ & .479 & $+.033^*$ & .872 & $+.000$ \\
\midrule
\multicolumn{5}{l}{\emph{Boxes}} \\
NMS, raw scores      & .429 & $-.017$   & \gry{.872} & \gry{$+.000$} \\
NMS, z-scored scores & .399 & $-.047^*$ & \gry{.872} & \gry{$+.000$} \\
\midrule
\multicolumn{5}{l}{\emph{Presence}} \\
Logit, $\omega=0.5$  & \gry{.479} & \gry{$+.033^*$} & .844 & $-.028^*$ \\
Mean probability     & \gry{.479} & \gry{$+.033^*$} & .821 & $-.051^*$ \\
Mean rank            & \gry{.479} & \gry{$+.033^*$} & .836 & $-.035^*$ \\
Maximum              & \gry{.479} & \gry{$+.033^*$} & .857 & $-.015^*$ \\
\bottomrule
\end{tabular}

\vspace{6pt}
\textbf{(c)} Learned fusion vs.\ late fusion (S@1)\\[2pt]
\begin{tabular}{l ccccc}
\toprule
 & & & \multicolumn{2}{c}{Gated, one expert} & \\
\cmidrule(lr){4-5}
Partner of DINOv3 & Concat & Gated & DINOv3 & partner & Late \\
\midrule
V-JEPA~2.1, image & .344 & .284 & .254 & .196 & \textbf{.479} \\
C-RADIOv4         & .317 & .192 & .088 & .188 & .442 \\
PE-Spatial        & .175 & .324 & .294 & .163 & .457 \\
\bottomrule
\end{tabular}
\end{minipage}
\end{table*}

\subsection{Late fusion with V-JEPA~2.1}
\label{sec:res-fusion}

Fusing the DINOv3 and V-JEPA~2.1 (image) heads raises S@1 from $0.446$ to
$0.479$ ($+0.033$, paired $[+0.002, +0.062]$), S@5 by $0.030$
($[+0.016, +0.044]$) and top-five mAP from $0.128$ to $0.154$, on easy and hard scenes alike (\cref{tab:errors}). Presence
AUROC stays at $0.872$ ($[-0.002, +0.002]$).

\paragraph{Why it helps.}
The heads fail on different queries. Of the $704$ present queries, V-JEPA~2.1
alone localizes $64$ that DINOv3 misses and DINOv3 alone $144$, so picking the
better head per query would reach S@1 $0.537$. Fusion recovers $42$ of the $64$
and fixes $29$ queries where neither top-1 box is correct but both heads place
boxes near the target. It loses $48$ of DINOv3's hits, $31$ by adopting
V-JEPA~2.1's wrong top-1 box (\cref{fig:fusion}): a net gain of $23$ queries.

\paragraph{Fusion rule.}
Both halves matter (\cref{tab:ablation}b). Non-maximum suppression keeps
whichever head's box scores higher and falls below DINOv3 alone, while weighted
box fusion rewards boxes that both heads propose. Every presence rule that
gives V-JEPA~2.1 more weight lowers AUROC, by $0.015$ to $0.051$, because
V-JEPA~2.1 ranks presence worse ($0.714$ alone).

\paragraph{What the gain depends on.}
Read as a two-frame clip, V-JEPA~2.1 adds nothing ($-0.003$,
$[-0.033, +0.026]$); PE-Spatial ($+0.011$) and C-RADIOv4 ($-0.004$) leave S@1
within the retraining spread, and DINOv2 lowers it ($-0.034$,
$[-0.061, -0.006]$). Partner accuracy alone does not explain this, since the
clip-read V-JEPA~2.1 ($0.290$) is stronger than the three image partners. Box
fusion needs agreement: PE-Spatial, whose per-query oracle ($0.531$) is close
to V-JEPA~2.1's ($0.537$), agrees with DINOv3 on far fewer correct boxes ($88$
queries against $170$). The image-read V-JEPA~2.1 is also the strongest
partner, so partner strength and representation are not fully separated.

\paragraph{Learned fusion.}
For every partner, a head trained on both backbones falls below both late
fusion and DINOv3 alone (\cref{tab:ablation}c). Inside the gated heads, the
DINOv3 expert alone reaches only $0.088$ to $0.294$: joint training weakens the
stronger backbone's path.

\subsection{What the head uses}
\label{sec:res-ablation}

Conditioning on another test object's exemplar collapses S@1 for every backbone
(\cref{tab:ablation}a), for DINOv3 to $0.054$ ($-0.392$, $[-0.431, -0.355]$):
each head localizes the object it is shown, not the most salient one. Removing
the correlation features separates the backbones. DINOv3 falls to $0.024$,
while PE-Spatial and C-RADIOv4 lose at most $0.013$, so their heads read the
exemplar through FiLM (\cref{eq:film}). DINOv3 is the backbone whose patch
similarities carry the instance signal, consistent with
\cref{sec:res-backbones}: heads that match through FiLM do well on seen objects
and poorly on new ones. Pooling the exemplar costs DINOv3 $0.202$ S@1, adding
cross-attention to the dense head costs $0.193$, and four exemplar views do not
help. V-JEPA~2.1 follows the same pattern (Supp.~D).

Zeroing the text embedding changes S@1 by at most $0.002$, as expected when
training uses one sentence for every object. Passing the object's name instead
raises PE-Spatial's S@1 by $0.040$ ($[+0.007, +0.075]$) and changes neither
C-RADIOv4 nor V-JEPA~2.1 reliably. The main model uses the generic sentence, so
we do not describe it as language-conditioned.

\begin{table}[t]
\centering\footnotesize
\caption{\textbf{Errors and cost.} \emph{Top}: S@1 on the easy and hard
scenes, and top-1 outcomes of all $3{,}200$ queries at $p \geq 0.5$: a present
query is a correct detection (TP), a mislocalization (FP-loc.) or a miss (FN);
an absent query is a false alarm (FP-abs.) or a correct rejection (TN).
\emph{Bottom}: trainable parameters (frozen encoders not counted), scene
encoding ($72$ regions), head time
per query, time per query amortized over a scene's $20$ queries, and peak GPU
memory. Baselines have no shared scene encoding; template matching runs on the
CPU. The fused heads run in turn, so their times add and peak memory is the
larger head's.}
\label{tab:errors}
\setlength{\tabcolsep}{2.5pt}
\begin{tabular}{l cc ccc cc}
\toprule
 & \multicolumn{2}{c}{S@1} & \multicolumn{3}{c}{Present} & \multicolumn{2}{c}{Absent} \\
\cmidrule(lr){2-3}\cmidrule(lr){4-6}\cmidrule(lr){7-8}
Model & easy & hard & TP & FP-loc. & FN & FP-abs. & TN \\
\midrule
DINOv3            & \underline{.508} & \underline{.292} & 310 & 359 & 35  & 1{,}221 & 1{,}275 \\
V-JEPA~2.1, image & .388 & .193 & 189 & 329 & 186 & 988     & 1{,}508 \\
V-JEPA~2.1, video & .327 & .198 & 195 & 451 & 58  & 1{,}606 & 890 \\
C-RADIOv4         & .291 & .163 & 131 & 316 & 257 & 961     & 1{,}535 \\
DINOv2            & .267 & .163 & 114 & 361 & 229 & 1{,}001 & 1{,}495 \\
PE-Spatial        & .233 & .153 & 76  & 306 & 322 & 851     & 1{,}645 \\
SigLIP~2          & .233 & .104 & 94  & 311 & 299 & 909     & 1{,}587 \\
\midrule
Late fusion (ours) & \textbf{.540} & \textbf{.327} & 330 & 339 & 35 & 1{,}194 & 1{,}302 \\
\bottomrule
\end{tabular}

\vspace{4pt}
\begin{tabular}{l ccccc}
\toprule
 & Params & Scene & Head & Query & GPU \\
Method & (M) & (s) & (ms) & (ms) & (GB) \\
\midrule
Template matching   & 0    & --   & --  & 150       & --   \\
OWLv2, image-guided & 0    & --   & --  & 35{,}960  & 0.8  \\
Qwen2.5-VL-7B       & 0    & --   & --  & 1{,}409   & 17.9 \\
\midrule
DINOv3 head         & 3.26 & 0.89 & 28  & 72        & 9.0  \\
V-JEPA~2.1 head     & 3.39 & 4.08 & 28  & 233       & 10.0 \\
Late fusion         & 6.65 & 4.97 & 56  & 305       & 10.0 \\
\bottomrule
\end{tabular}
\end{table}

\subsection{Errors and cost}
\label{sec:res-other}

\paragraph{Errors.}
Presence ranks queries well but is not calibrated. At $p \geq 0.5$, false
alarms on absent queries are the most frequent error of every model
(\cref{tab:errors}): $1{,}194$ of $2{,}496$ for the fused model. Scene presence
is the maximum over $72$ regions, so one confident region is enough for a false
alarm. On present queries, DINOv3 and the fused model rarely miss outright
($35$ each); most of their errors are boxes in the wrong place.

\paragraph{Cost.}
With scene encoding shared by a scene's $20$ queries, the DINOv3 head needs
$72$\,ms per query. Late fusion raises scene encoding from $0.89$ to $4.97$\,s,
about $5.6\times$, and the time per query to $305$\,ms. Qwen2.5-VL-7B needs
$1.41$\,s per query, and OWLv2, which our implementation runs anew on all $72$
regions for every query, $36.0$\,s.

\section{Conclusion}
\label{sec:conclusion}

WALDO localizes a specific, previously unseen object instance from one
reference photograph and predicts whether it is in the scene, using a small
head on frozen features and no per-object adaptation. The head localizes the
object it is shown rather than another one, and for DINOv3, dense correlation
between scene patches and exemplar tokens carries this signal. Features of a video world model,
V-JEPA~2.1, complement those of an image model when they are read through the
model's image pathway and combined with DINOv3 at the output rather than
inside one head.

\paragraph{Limitations.}
Each arm is trained once, and the fusion gain ($0.033$) is about the size of
the retraining spread ($0.029$); its paired interval excludes zero, but repeated
seeds are needed to confirm it. The fused model runs two backbones. The heads
are trained only on synthetic composites, S@1 drops from $0.540$ on easy to
$0.327$ on hard scenes, and presence needs a calibrated threshold. On a second benchmark of cars and aircraft (Supp.~F), where the exemplar is
cropped from the searched scene, removing correlation or swapping the exemplar
changes no head's S@1 significantly, so the heads match the category, not the
instance. DINOv3, whose InsDet lead comes from correlation, falls there below
V-JEPA~2.1 read as a clip, and late fusion with V-JEPA~2.1 does not help, so
our backbone ranking and fusion gain are shown on InsDet only.

\section*{LLM Usage Statement}
\label{sec:llmusagestatement}
We used LLM-based coding and writing assistants. ChatGPT (GPT-5) was used to
refine grammar, format the LaTeX of equations and tables, and condense some
sections to fit the page limit. Claude Code was used to write code for the
experiments and analysis, to check the text against the results, and to
revise the text for consistency. All experiments were run, and all results,
claims and citations were checked, by the authors, who take full
responsibility for the content.

{
    \small
    \bibliographystyle{ieeenat_fullname}
    \bibliography{main}
}

\clearpage
\appendix

\renewcommand{\thetable}{S\arabic{table}}
\renewcommand{\thefigure}{S\arabic{figure}}
\renewcommand{\theequation}{S\arabic{equation}}
\setcounter{table}{0}
\setcounter{figure}{0}
\setcounter{equation}{0}

\renewcommand{\topfraction}{0.9}
\renewcommand{\bottomfraction}{0.8}
\renewcommand{\textfraction}{0.1}
\renewcommand{\floatpagefraction}{0.7}
\renewcommand{\dbltopfraction}{0.9}
\renewcommand{\dblfloatpagefraction}{0.7}
\setcounter{topnumber}{4}
\setcounter{bottomnumber}{3}
\setcounter{totalnumber}{6}
\setcounter{dbltopnumber}{3}
\raggedbottom

%
\maketitlesupplementary
\renewcommand{\thesection}{\Alph{section}}
\setcounter{section}{0}

\section{Additional InsDet results}
\label{supp:insdet}

\paragraph{Closed set.}
\Cref{tab:supp-closed} trains each head on composites of all $100$ objects, so
every queried object is also a training object. Scored on the same $20$ test
objects as the novel-object protocol, the closed-set heads reach similar S@1,
and novelty costs DINOv3 far less than PE-Spatial or C-RADIOv4.

\begin{table}[t]
\centering\footnotesize
\caption{\textbf{Closed-set reference} (S@1, view-$0$ exemplar). Closed: heads
trained on all $100$ objects. Novel: the main-paper heads, trained without the
$20$ test objects. DINOv3 was trained twice in the closed set with the same
configuration.}
\label{tab:supp-closed}
\setlength{\tabcolsep}{4pt}
\begin{tabular}{l ccc}
\toprule
 & Closed, & Closed, & Novel, \\
Backbone & $100$ objects & $20$ test objects & $20$ test objects \\
\midrule
DINOv3 (two runs)   & .528 / .500 & .534 / .510 & .446 \\
PE-Spatial          & .533 & .501 & .210 \\
C-RADIOv4           & .548 & .528 & .254 \\
V-JEPA~2.1, video   & .421 & .430 & .290 \\
\bottomrule
\end{tabular}
\end{table}

\paragraph{Validation objects.}
\Cref{tab:supp-val} scores the same heads on the $10$ validation objects, each
queried in all $160$ real scenes ($1{,}600$ queries, $346$ present). These
queries chose only the fusion presence weight (\cref{supp:fusion}), never a
checkpoint, so they are a second held-out object set. DINOv3 again has the
best presence AUROC and mAP, but its S@1 lead does not hold: the V-JEPA~2.1
image head reaches $0.341$ against DINOv3's $0.315$, with overlapping
intervals. With $10$ and $20$ objects, the S@1 ranking of the two strongest
backbones depends on which objects are queried, while DINOv3's presence lead
holds on both object sets.

\begin{table}[t]
\centering\footnotesize
\caption{\textbf{Validation objects} ($10$ objects, $1{,}600$ queries). The
last column repeats the test S@1 of the same checkpoint (main paper, Tab.~2).
\textbf{Bold}: best per column.}
\label{tab:supp-val}
\setlength{\tabcolsep}{3pt}
\begin{tabular}{l cccccc}
\toprule
 & \multicolumn{5}{c}{Validation objects} & Test \\
\cmidrule(lr){2-6}
Backbone & S@1 & 95\% CI & S@5 & AUROC & mAP & S@1 \\
\midrule
DINOv3              & .315 & [.271, .361] & \textbf{.569} & \textbf{.817} & \textbf{.089} & \textbf{.446} \\
PE-Spatial          & .246 & [.211, .284] & .523 & .660 & .025 & .210 \\
C-RADIOv4           & .283 & [.239, .330] & .564 & .694 & .041 & .254 \\
V-JEPA~2.1, video   & .246 & [.206, .285] & .509 & .690 & .041 & .290 \\
V-JEPA~2.1, image   & \textbf{.341} & [.302, .385] & .540 & .701 & .077 & .332 \\
\bottomrule
\end{tabular}
\end{table}

\section{Late-fusion analysis}
\label{supp:fusion}

\paragraph{Per object.}
\Cref{tab:supp-perobj} breaks the fused S@1 of the main paper ($0.479$
against $0.446$) down by test object. Fusion raises S@1 on $11$ of the $20$
objects, lowers it on $6$ and leaves $3$ unchanged. The largest gains come
where V-JEPA~2.1 is at least as accurate as DINOv3 (Costa caramel,
$0.692 \to 0.923$; thermos flask, $0.724 \to 0.931$), and most losses where
it is far weaker (Japanese chocolate, $0.500 \to 0.364$, with V-JEPA~2.1 at
$0.045$). The gain is therefore not carried by one or two objects, but a
partner that is much weaker on an object still costs accuracy on it.

\begin{table}[t]
\centering\footnotesize
\caption{\textbf{S@1 per test object}, present queries only ($n$ per object),
sorted by the fused S@1. Late fusion: weighted box fusion of each head's top
five boxes, as in the main paper.}
\label{tab:supp-perobj}
\setlength{\tabcolsep}{4pt}
\begin{tabular}{l rccc}
\toprule
Object & $n$ & DINOv3 & V-JEPA~2.1 & Late fusion \\
\midrule
Thermos flask purple    & 29 & .724 & .828 & .931 \\
Costa caramel           & 52 & .692 & .750 & .923 \\
Essential oil bergamot  & 41 & .805 & .463 & .854 \\
Skinceuticals BE        & 11 & .818 & .182 & .727 \\
Ousa grated cheese      & 16 & .688 & .375 & .688 \\
Japanese finger cookies & 26 & .500 & .462 & .654 \\
Acnes cream             & 15 & .667 & .133 & .600 \\
Binder clips median     & 54 & .593 & .333 & .519 \\
Binder clips small      & 50 & .420 & .460 & .500 \\
Baseball cap black      & 33 & .636 & .424 & .485 \\
HB marine collagen      & 47 & .447 & .277 & .468 \\
Lindor salted caramel   & 75 & .440 & .387 & .440 \\
Japanese chocolate      & 22 & .500 & .045 & .364 \\
Kaneyo kitchen bleach   & 22 & .273 & .182 & .364 \\
HB calcium              & 51 & .196 & .176 & .255 \\
Banana milk drink       & 37 & .189 & .216 & .216 \\
Watercolor penbox       & 52 & .173 & .077 & .212 \\
Geo roast charcoal      & 27 & .111 & .222 & .185 \\
Handcream rosewater     & 19 & .158 & .053 & .158 \\
Instant noodle black    & 25 & .160 & .000 & .080 \\
\midrule
All                     & 704 & .446 & .332 & .479 \\
\bottomrule
\end{tabular}
\end{table}

\paragraph{Presence weight.}
The presence logit weight $\omega$ on DINOv3 is the only fitted parameter of
late fusion. It is chosen on the validation objects from a grid of step $0.05$
by presence AUROC (\cref{tab:supp-omega}). The curve is flat above
$\omega=0.75$: the best weight, $0.9$, gains only $0.002$ over DINOv3 alone on
validation and leaves test AUROC at $0.872$. Presence fusion therefore keeps
DINOv3's presence score almost unchanged, and the S@1 gain of late fusion comes
from the boxes, which $\omega$ does not affect.

\begin{table}[t]
\centering\footnotesize
\caption{\textbf{Presence AUROC on the validation objects} against the logit
weight $\omega$ on DINOv3 ($1{,}600$ queries). $\omega=0$: V-JEPA~2.1 alone;
$\omega=1$: DINOv3 alone. \textbf{Bold}: chosen weight, the highest AUROC
before rounding.}
\label{tab:supp-omega}
\setlength{\tabcolsep}{3.5pt}
\begin{tabular}{l cccccccc}
\toprule
$\omega$ & 0 & 0.5 & 0.7 & 0.8 & 0.85 & 0.9 & 0.95 & 1 \\
\midrule
AUROC & .701 & .793 & .815 & .819 & .820 & \textbf{.820} & .819 & .817 \\
\bottomrule
\end{tabular}
\end{table}

\section{Learned fusion}
\label{supp:learned}

Sec.~5.2 of the main paper reports that heads trained on two backbones fall
below DINOv3 alone on the test scenes. \Cref{tab:supp-learned} adds the
synthetic validation score on which every checkpoint was selected. On
synthetic composites the gated heads reach $0.722$ to $0.766$ S@1, level with
the DINOv3 head ($0.771$); on real scenes they fall to $0.192$ to $0.324$,
against $0.446$. Concatenation behaves the same way. Learned fusion thus fits
the synthetic episodes as well as one backbone does, but transfers worse to
real clutter, and selection on synthetic validation cannot detect it. Late
fusion has nothing to fit on the synthetic data and does not show this drop.

\begin{table}[t]
\centering\footnotesize
\caption{\textbf{Learned fusion: synthetic validation vs.\ real test.}
Synthetic val.: S@1 of the selected checkpoint on the $400$ fixed validation
composites. The DINOv3 run predates the episode-seeding fix of
\cref{supp:impl} and was validated on a different draw of the $400$ composites
($271$ present targets instead of $248$). Expert-only scores of the gated
heads are in Tab.~3c of the main paper.}
\label{tab:supp-learned}
\setlength{\tabcolsep}{4pt}
\begin{tabular}{l ccc}
\toprule
 & Synthetic val. & \multicolumn{2}{c}{Real test} \\
\cmidrule(lr){3-4}
Head & S@1 & S@1 & AUROC \\
\midrule
DINOv3 alone              & .771 & .446 & .872 \\
\midrule
V-JEPA~2.1 (image) alone  & .734 & .332 & .714 \\
\quad + DINOv3, concat    & .710 & .344 & .723 \\
\quad + DINOv3, gated     & .750 & .284 & .776 \\
\midrule
PE-Spatial alone          & .637 & .210 & .639 \\
\quad + DINOv3, concat    & .710 & .175 & .673 \\
\quad + DINOv3, gated     & .766 & .324 & .641 \\
\midrule
C-RADIOv4 alone           & .758 & .254 & .681 \\
\quad + DINOv3, concat    & .770 & .317 & .742 \\
\quad + DINOv3, gated     & .722 & .192 & .657 \\
\bottomrule
\end{tabular}
\end{table}

\section{Ablations on the other backbones}
\label{supp:ablations}

\paragraph{V-JEPA~2.1.}
\Cref{tab:supp-vjepa} repeats the ablations of Sec.~5.3 of the main paper on
the V-JEPA~2.1 head read as a two-frame clip. Its only reliable gain is the
image pathway, which costs AUROC ($-0.028$, $[-0.046, -0.010]$).

\begin{table}[t]
\centering\footnotesize
\caption{\textbf{V-JEPA~2.1 input pathway and head.} $\Delta$ against the
two-frame video head; $^*$paired 95\% interval excludes zero.}
\label{tab:supp-vjepa}
\begin{tabular}{l ccc}
\toprule
Variant & S@1 & $\Delta$S@1 & AUROC \\
\midrule
Two-frame video (reference) & .290 & ref.      & .742 \\
\midrule
Image pathway               & .332 & $+.043^*$ & .714 \\
4-view exemplar             & .277 & $-.013^*$ & .740 \\
Pooled exemplar             & .217 & $-.072^*$ & .649 \\
Dense + cross-attention     & .317 & $+.027$   & .662 \\
Object name as text         & .273 & $-.017$   & .666 \\
\bottomrule
\end{tabular}
\end{table}

\paragraph{Head-free feature probe.}
To separate the backbone from the trained head, we score the frozen features
with no training at all (\cref{tab:supp-probe}). For every query, each scene
patch of the $72$ regions is compared by cosine similarity with the exemplar's
object tokens, either with the best-matching exemplar token (max) or with
their masked mean (pooled). \emph{Point@1} counts present queries whose
most similar patch lies inside a ground-truth box; \emph{Point@5} accepts any
of five distinct peaks; AUROC uses the scene's highest similarity as the
presence score. A random point lands in a box with probability $0.002$.

The probe ranks DINOv3 first, as the trained heads do, but it places
C-RADIOv4 close behind ($0.571$ against $0.652$ point@1) although its head
reaches only $0.254$ S@1. C-RADIOv4's features match novel exemplars almost as
well as DINOv3's; the gap opens in training, consistent with the closed-set
result of \cref{tab:supp-closed}, where the C-RADIOv4 head reaches $0.528$ on
the same objects once it has seen them. V-JEPA~2.1 is weakest head-free
($0.21$ to $0.23$) yet its trained head beats PE-Spatial and C-RADIOv4, and
the image pathway that helps its head ($+0.043$) barely changes the raw
similarity ($0.207$ against $0.226$ point@1). Earlier blocks match far worse for every backbone (DINOv3 block
$8$: $0.145$), except PE-Spatial, whose block $8$ ($0.358$) is as good as its
last; averaging its four blocks reaches $0.422$, yet the trained four-block
head is worse (main paper, Tab.~3a). Point@$K$ asks for one patch inside the
box and S@1 for a box with IoU $\geq 0.5$, so the two are compared by rank,
not by value.

\begin{table}[t]
\centering\footnotesize
\caption{\textbf{Head-free probe} of the frozen features on the $20$ test
objects ($3{,}200$ queries, $704$ present; last block, view-$0$ exemplar). No
parameter is trained. Head S@1: the trained head on the same features (main
paper). Chance point@1: $0.002$.}
\label{tab:supp-probe}
\setlength{\tabcolsep}{2.5pt}
\begin{tabular}{l c ccc c}
\toprule
 & Max & \multicolumn{3}{c}{Pooled exemplar} & Head \\
\cmidrule(lr){2-2}\cmidrule(lr){3-5}
Features & Pt@1 & Pt@1 & Pt@5 & AUROC & S@1 \\
\midrule
DINOv3                      & \textbf{.608} & \textbf{.652} & \textbf{.878} & \textbf{.871} & \textbf{.446} \\
C-RADIOv4                   & .430 & .571 & .861 & .848 & .254 \\
C-RADIOv4, $512$\,px pooled & .499 & .572 & .851 & .862 & .266 \\
PE-Spatial                  & .162 & .334 & .611 & .724 & .210 \\
PE-Spatial, $512$\,px pooled & .165 & .355 & .661 & .693 & .180 \\
V-JEPA~2.1, 2-frame clip    & .145 & .226 & .415 & .631 & .290 \\
V-JEPA~2.1, image           & .179 & .207 & .331 & .616 & .332 \\
V-JEPA~2.1, 16-frame clip   & .175 & .233 & .425 & .643 & --   \\
\bottomrule
\end{tabular}
\end{table}

\section{Qualitative results}
\label{supp:qualitative}

\begin{figure*}[t]
\centering
\includegraphics[width=\textwidth]{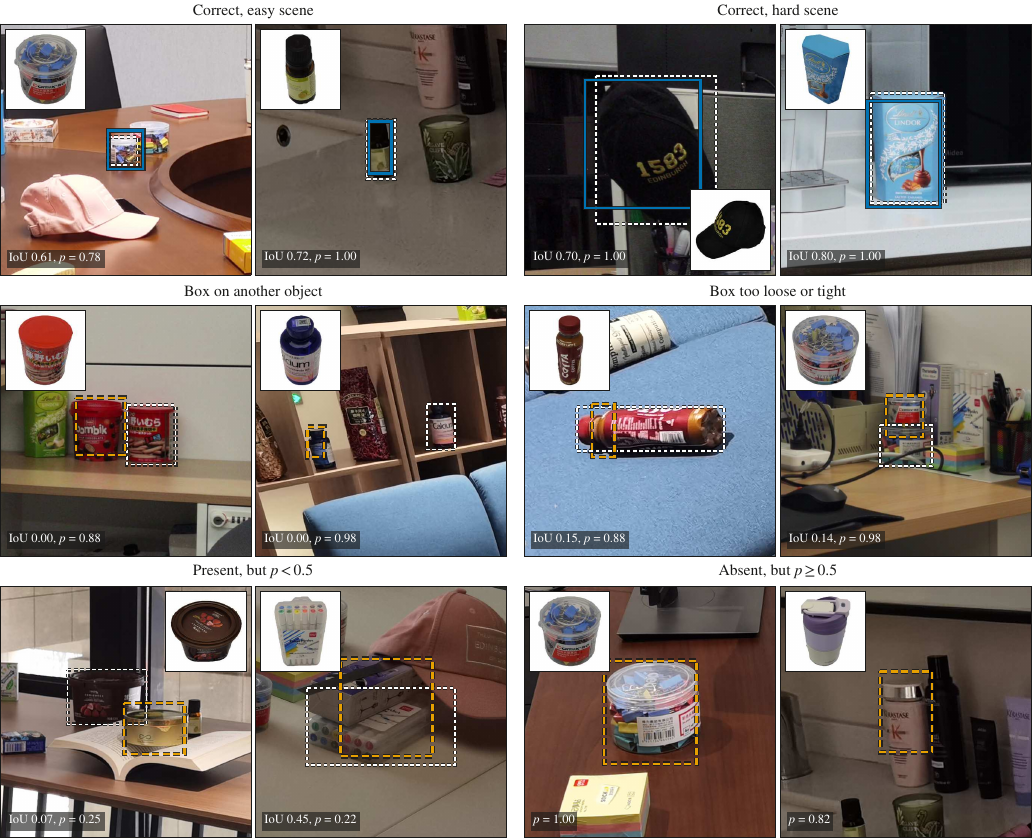}
\caption{\textbf{Late fusion on novel test objects}, two random queries per
outcome group. Inset: the view-$0$ exemplar given to the model. White dashed:
ground truth. Blue solid: correct top-1 box (IoU $\geq 0.5$); orange dashed:
wrong top-1 box. $p$: fused presence probability. Group sizes over the
$3{,}200$ test queries: $264$ correct in easy scenes, $66$ correct in hard
scenes, $247$ top-1 boxes on another object, $56$ with $0.1 \leq$ IoU $< 0.5$,
$35$ present queries with $p<0.5$, and $1{,}194$ absent queries with
$p \geq 0.5$.}
\label{fig:supp-qualitative}
\end{figure*}

\Cref{fig:supp-qualitative} shows the fused model on test queries from six
outcome groups. The examples are drawn at random (seed~$0$) within each group,
among queries whose boxes fit in one readable crop, so they are not selected
by hand. In the failures shown, the wrong top-1 boxes and the false alarms
land on objects that share the exemplar's shape or colour, such as another red
can or a larger jar of binder clips, and the loose boxes cover only part of
the object.
The presence probability is not calibrated: $1{,}194$ of the $2{,}496$ absent
queries receive $p \geq 0.5$, while $35$ of the $704$ present ones receive
$p < 0.5$. AUROC depends only on the ranking of $p$, but a deployed system
would need a threshold chosen on validation data.

\section{Car and aircraft benchmark}
\label{supp:cardata}

We also evaluate on a set of $220$ scenes of dense parking lots and aircraft
bays ($103$ car, $117$ aircraft) at $512\times512$\,px with $763$ instance
boxes, split by scene at seed~$0$ into $150$ training, $35$ validation and $35$
test scenes. It tests a different regime: many near-identical instances of one
category per scene, a single forward pass over the whole image, and exemplars
cropped from the scene in which the instance must be found, which makes it
easier than the cross-view setting of InsDet. A \emph{catalogue} query asks
for any instance of the exemplar's category ($35$ queries), an
\emph{instance} query for the cropped instance itself ($116$ queries). Hard
absence episodes are sampled so that object count does not predict the label:
count alone reaches AUROC $0.500$ on them. The same backbones, head variants
and fusions as on InsDet are trained on this set (\cref{tab:cardata}).

The InsDet ranking does not replicate here. DINOv3 is near the bottom
($0.172$ instance S@1), cross-attention raises it to $0.362$, V-JEPA~2.1 is
better read as a clip than as an image, and with the image-read V-JEPA~2.1 a
gated head ($0.267$) beats late fusion ($0.181$; \cref{tab:cardata-fusion}). Late fusion
with the presence weight fixed from InsDet helps significantly for one partner
only, SigLIP~2 ($+0.052$). The exemplar is cropped from the scene itself, so
template matching reaches $1.000$, and no head uses the exemplar's identity
(\cref{tab:cardata-abl}). Correlation features can be removed and the exemplar
swapped for another instance with no significant change in instance S@1 for
any backbone. Zeroing the exemplar changes mainly hard-absence AUROC, so the
exemplar tells the heads which category is asked for. On InsDet, removing
correlation collapses DINOv3 ($0.446 \to 0.024$, main paper, Tab.~3a); here
it changes nothing, so the patch matching that gives DINOv3 its InsDet lead is
not used. We therefore read this set as a test of category-level matching, not
as a replication of the InsDet results.

\begin{figure*}[t]
\centering
\includegraphics[width=\textwidth]{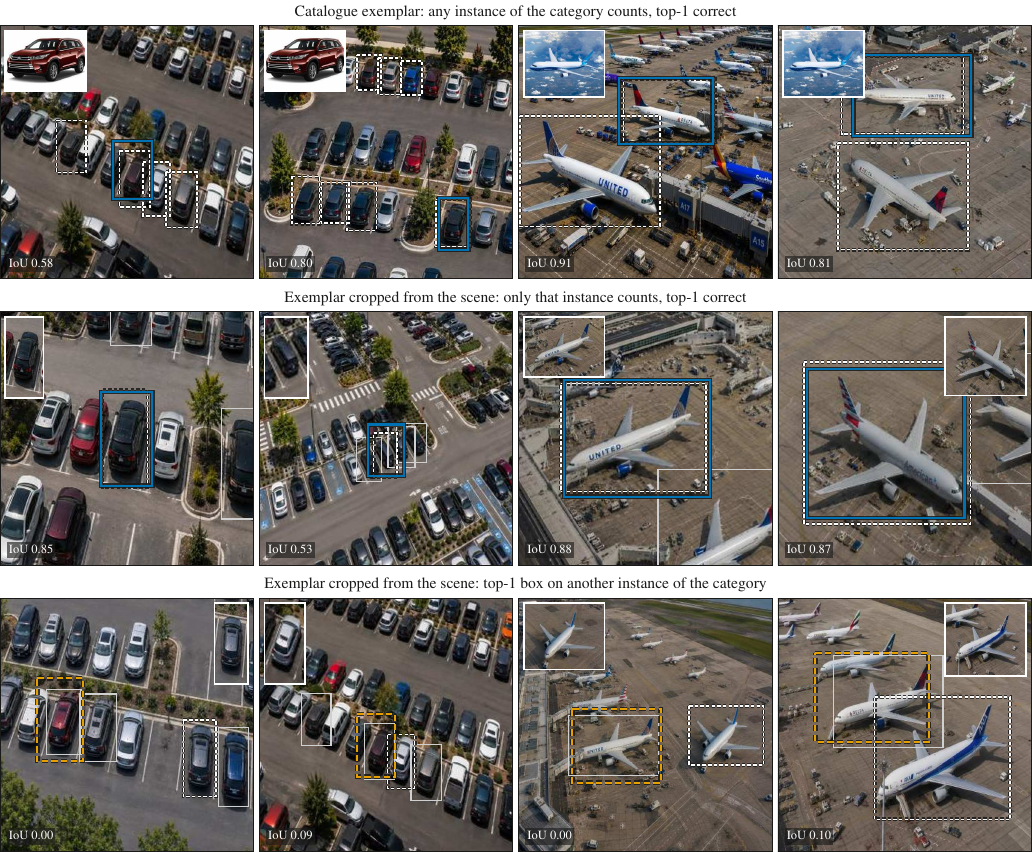}
\caption{\textbf{Cars and aircraft}, DINOv3 with cross-attention, test split.
Four random queries per row (two car and two aircraft scenes), cropped around
the boxes. Inset: the exemplar given to the head, either the catalogue photo
(top row) or the target cropped from the scene with $15\%$ padding. White
dashed: the instances that count as correct; thin grey: the other labelled
instances. Blue solid: correct top-1 box (IoU $\geq 0.5$); orange dashed:
wrong top-1 box. Over the test split, $25$ of $35$ catalogue queries and $42$
of $116$ instance queries are correct, and $42$ instance queries put the top-1
box on another labelled instance.}
\label{fig:supp-cardata}
\end{figure*}

\Cref{fig:supp-cardata} shows the best head on this set, DINOv3 with
cross-attention, on random test queries. It finds the category reliably ($25$
of $35$ catalogue queries correct), but of the $74$ instance queries it gets
wrong, $42$ place the top-1 box on another labelled vehicle or aircraft of the
same category. This is the failure the ablations predict: the head answers
which category is asked for, not which instance. The remaining $32$ boxes
overlap no labelled instance. Only $3.5$ instances per scene are labelled on
average, so some of these may sit on unlabelled vehicles.

\begin{table*}[t]
\centering\footnotesize
\caption{\textbf{Cars and aircraft}, test split: every trained head, sorted by
instance S@1. Intervals: $95\%$ scene bootstrap. Hard abs.: presence AUROC on
count-matched absence episodes (count alone: $0.500$). Template matching slides
the exemplar crop over the scene it was cut from.}
\label{tab:cardata}
\setlength{\tabcolsep}{5pt}
\begin{tabular}{l c ccc c}
\toprule
 & Catalogue & \multicolumn{3}{c}{Instance} & Hard abs. \\
\cmidrule(lr){3-5}
Head & S@1 & S@1 & 95\% CI & S@5 & AUROC \\
\midrule
Template matching                     & --   & 1.000 & -- & -- & -- \\
\midrule
DINOv3 (cross-attn.)                  & .714 & .362 & [.264, .481] & .629 & .943 \\
DINOv3 + V-JEPA~2.1, gated (image)    & .686 & .267 & [.190, .364] & .552 & .929 \\
V-JEPA~2.1                            & .629 & .250 & [.179, .346] & .534 & .904 \\
DINOv3 + PE-Spatial, gated            & .657 & .233 & [.164, .323] & .534 & .873 \\
V-JEPA~2.1 (cross-attn.)              & .657 & .233 & [.167, .319] & .526 & .865 \\
SigLIP~2                              & .657 & .224 & [.156, .302] & .448 & .724 \\
V-JEPA~2.1 (pooled exemplar)          & .514 & .215 & [.149, .301] & .457 & .869 \\
C-RADIOv4 (cross-attn.)               & .657 & .215 & [.144, .302] & .552 & .816 \\
PE-Spatial                            & .714 & .207 & [.143, .277] & .543 & .847 \\
C-RADIOv4 ($512$\,px pooled)          & .686 & .207 & [.147, .281] & .526 & .811 \\
PE-Spatial (pooled exemplar)          & .686 & .198 & [.136, .269] & .483 & .828 \\
DINOv3 + V-JEPA~2.1, gated            & .686 & .198 & [.129, .284] & .543 & .858 \\
DINOv3 + V-JEPA~2.1, concat           & .657 & .198 & [.128, .283] & .483 & .837 \\
DINOv3 + V-JEPA~2.1, concat (image)   & .629 & .190 & [.128, .260] & .457 & .759 \\
DINOv3 + C-RADIOv4, concat            & .657 & .190 & [.122, .269] & .457 & .834 \\
DINOv3 + PE-Spatial, concat           & .714 & .190 & [.129, .261] & .534 & .847 \\
PE-Spatial (cross-attn.)              & .571 & .190 & [.121, .272] & .474 & .843 \\
PE-Spatial (four blocks)              & .629 & .190 & [.128, .260] & .474 & .795 \\
C-RADIOv4 (four blocks)               & .714 & .190 & [.123, .264] & .500 & .849 \\
C-RADIOv4                             & .657 & .181 & [.121, .253] & .491 & .705 \\
DINOv3 + C-RADIOv4, gated             & .600 & .181 & [.119, .255] & .491 & .821 \\
C-RADIOv4 (pooled exemplar)           & .571 & .181 & [.115, .265] & .483 & .603 \\
DINOv3                                & .629 & .172 & [.106, .253] & .422 & .779 \\
DINOv3 (pooled exemplar)              & .571 & .172 & [.105, .247] & .491 & .672 \\
V-JEPA~2.1 (image)                    & .543 & .164 & [.103, .240] & .405 & .771 \\
DINOv2                                & .543 & .155 & [.092, .226] & .448 & .807 \\
PE-Spatial ($512$\,px pooled)         & .543 & .155 & [.093, .228] & .397 & .789 \\
\bottomrule
\end{tabular}
\end{table*}

\begin{table*}[t]
\begin{minipage}[t]{0.49\textwidth}
\centering\footnotesize
\caption{\textbf{Late fusion on cars and aircraft}: DINOv3 with each partner,
the InsDet rule with the presence weight fixed at $0.9$ (not tuned on this
set). $\Delta$: fused minus DINOv3 alone, instance S@1, with paired $95\%$
interval. DINOv3 alone: $.629$ / $.172$ / $.422$ / $.779$.}
\label{tab:cardata-fusion}
\setlength{\tabcolsep}{2.5pt}
\begin{tabular}{l cccc c}
\toprule
 & Cat. & \multicolumn{3}{c}{Instance} & Hard \\
\cmidrule(lr){3-5}
Partner & S@1 & S@1 & $\Delta$S@1 & S@5 & AUROC \\
\midrule
V-JEPA~2.1, image & .600 & .181 & $+.009$ [$.000$, $+.029$] & .431 & .785 \\
V-JEPA~2.1, video & .686 & .207 & $+.035$ [$-.009$, $+.083$] & .474 & .853 \\
SigLIP~2          & .686 & .224 & $+.052$ [$+.015$, $+.104$] & .474 & .776 \\
PE-Spatial        & .629 & .190 & $+.017$ [$-.029$, $+.066$] & .474 & .819 \\
C-RADIOv4         & .686 & .172 & $\phantom{+}.000$ [$-.026$, $+.023$] & .457 & .769 \\
DINOv2            & .600 & .164 & $-.009$ [$-.028$, $.000$] & .431 & .794 \\
\bottomrule
\end{tabular}
\end{minipage}
\hfill
\begin{minipage}[t]{0.49\textwidth}
\centering\footnotesize
\caption{\textbf{Inference-time ablations on cars and aircraft}, same
checkpoint. $\Delta$ instance S@1 for each change, and $\Delta$ hard-absence
AUROC for the zeroed exemplar. Swapped exemplar: another instance's crop from
the same scene (single-instance scenes skipped). $^*$Paired $95\%$ interval
excludes zero.}
\label{tab:cardata-abl}
\setlength{\tabcolsep}{2.5pt}
\begin{tabular}{l cccc c}
\toprule
 & \multicolumn{4}{c}{$\Delta$ instance S@1} & $\Delta$AUROC \\
\cmidrule(lr){2-5}\cmidrule(lr){6-6}
Backbone & No corr. & Swap ex. & Swap text & Zero ex. & Zero ex. \\
\midrule
DINOv3              & $+.000$ & $+.000$ & $+.000$ & $-.009$   & $-.114^*$ \\
DINOv2              & $+.000$ & $+.009$ & $-.009$ & $+.000$   & $-.384^*$ \\
SigLIP~2            & $+.000$ & $-.018$ & $-.009$ & $-.026$   & $-.089^*$ \\
PE-Spatial          & $+.000$ & $-.018$ & $+.009$ & $+.000$   & $-.202^*$ \\
C-RADIOv4           & $+.000$ & $+.009$ & $+.000$ & $-.017$   & $-.344^*$ \\
V-JEPA~2.1, video   & $+.000$ & $-.044$ & $-.009$ & $-.086^*$ & $-.188^*$ \\
V-JEPA~2.1, image   & $+.000$ & $+.000$ & $+.000$ & $+.000$   & $-.090^*$ \\
\bottomrule
\end{tabular}
\end{minipage}
\end{table*}

\section{Training and implementation details}
\label{supp:impl}

\paragraph{Data.}
One InsDet training object has no masks and is never pasted. The $400$
validation composites contain training and validation objects and are queried
only about the validation objects.

\paragraph{Exemplar augmentation.}
Exemplars are re-augmented every epoch: rotation of up to $\pm10^\circ$
(probability $0.5$), brightness and contrast jitter of $\times0.75$ to $1.25$
(probability $0.7$ each), saturation jitter of $\times0.8$ to $1.2$
(probability $0.5$), and Gaussian blur of radius $0.3$ to $1.2$\,px
(probability $0.25$). Padding the cut-out to a square with transparent pixels
keeps its aspect ratio and marks the padding as not-object.

\paragraph{Head.}
Class, summary and register tokens are discarded. The transformer blocks use
MLP ratio $3$ and dropout $0.1$. The head has $3.26$M parameters on a
$768$-dimensional backbone, $3.39$M on a $1024$-dimensional one and $3.46$M on
a $1152$-dimensional one; the token width changes only the input projections.
In the gated fusion head, the projected exemplars of the two backbones are
averaged before FiLM, and the gate's last layer is zero-initialized.

\paragraph{Optimization.}
The focal loss uses balance $0.25$ and focusing parameter $2$. The objectness
bias is initialized to $-\log 99$, so the background patches, about $99\%$ of
the grid, do not dominate the early gradients. AdamW uses weight decay $0.05$
and gradient clipping at $1.0$; the one-cycle schedule warms up over $5\%$ of
training, and training stops early after $12$ epochs without improvement.
Because the backbone is frozen, scene tokens and text embeddings are cached
after their first computation. Exemplars are not cached, so their augmentation
changes every epoch.

\paragraph{Episode seeding.}
Each episode is generated from a random state seeded by the run seed, the
episode index and the epoch. In the earliest runs this seed went through
Python's hashing of a tuple containing a string, which is randomized per
process. Those runs (the DINOv3 and SigLIP~2 heads, and the V-JEPA~2.1 heads
read as a clip: default, pooled exemplar, cross-attention, object name and
closed set) therefore drew their training and validation episodes from the
same generator and distribution, but not the identical episodes of the later
runs. This adds run-to-run variation of the kind measured by the retraining
spread in the main paper, not a systematic difference.

\paragraph{Backbones.}
The checkpoints are DINOv3~\cite{simeoni2025dinov3} ViT-B/16 (LVD-1689M),
DINOv2~\cite{oquab2023dinov2} ViT-B/14 (at $336$\,px, so that its patch of $14$
gives a $24\times24$ grid), SigLIP~2~\cite{tschannen2025siglip} SO400M/16 at
$384$\,px, PE-Spatial~\cite{bolya2026perception} B/16 (pretrained at
$512$\,px, run at $384$\,px), C-RADIOv4~\cite{ranzinger2026c} SO400M, and
V-JEPA~2.1~\cite{mur2026v} ViT-L/16 at $384$\,px with its predictor removed.
The $512$\,px variants of PE-Spatial and C-RADIOv4 pool their $32\times32$
grid to $24\times24$.

\paragraph{Implementation.}
The head and the evaluation are implemented in PyTorch without a detection
framework. V-JEPA~2.1 runs under bf16 autocasting; the other encoders and the
head run in fp32. Runtimes are measured with CUDA synchronization on one
NVIDIA RTX 6000 Ada Generation GPU ($48$\,GB): DINOv3 encodes a scene's $72$
regions in $0.89$\,s, shared by all queries on that scene, and the head adds
$28$\,ms per query. Training one head takes from $68$ minutes (DINOv3) to $247$
minutes (V-JEPA~2.1, image pathway) for at most $30$ epochs. A suite of $253$
unit tests, including the label-independence test of Sec.~3.2 of the main
paper, runs before training.

\end{document}